\documentclass{article} % For LaTeX2e
\usepackage{iclr2023_conference,times}

\usepackage{amsmath,amsfonts,bm}

\def\eqref#1{equation~\ref{#1}}
\def\1{\bm{1}}

\DeclareMathAlphabet{\mathsfit}{\encodingdefault}{\sfdefault}{m}{sl}
\SetMathAlphabet{\mathsfit}{bold}{\encodingdefault}{\sfdefault}{bx}{n}

\usepackage{hyperref}
\usepackage{url}

\usepackage{graphicx}
\usepackage[ruled,vlined]{algorithm2e}
\usepackage{subcaption}
\usepackage{wrapfig}

\title{Importance Estimation with Random Gradient for Neural Network Pruning}

\iclrfinalcopy

\author{Suman Sapkota \thanks{https://tsumansapkota.github.io/} \\
NepAl Applied Mathematics and Informatics Institute for Research (NAAMII), Nepal\\
\texttt{suman.sapkota@naamii.org.np} \\
\And
Binod Bhattarai \\
University of Aberdeen, UK \\
\texttt{binod.bhattarai@abdn.ac.uk} \\
}

\begin{document}

\maketitle

\begin{abstract}
Global Neuron Importance Estimation is used to prune neural networks for efficiency reasons.
To determine the global importance of each neuron or convolutional kernel, most of the existing methods either use activation or gradient information or both, which demands abundant labelled examples. In this work, we use heuristics to derive importance estimation similar to Taylor First Order (TaylorFO) approximation based methods. We name our methods \emph{TaylorFO-abs} and \emph{TaylorFO-sq}.
We propose two additional methods to improve these importance estimation methods. Firstly, we propagate \emph{random gradients} from the last layer of a network, thus avoiding the need for labelled examples. Secondly, we \emph{normalize} the gradient magnitude of the last layer output before propagating, which allows all examples to contribute similarly to the importance score. Our methods with additional techniques perform better than previous methods when tested on ResNet and VGG architectures on CIFAR-100 and STL-10 datasets. Furthermore, our method also complements the existing methods and improves their performances when combined with them. 

\end{abstract}

\section{Introduction and Background}

Neural Network Pruning~\cite{lecun1989optimal, han2015learning, gale2019state, blalock2020state} is one of the methods to reduce the parameters, compute and memory requirement. This method differs significantly from knowledge distillation~\cite{hinton2015distilling, gou2021knowledge} where a small model is trained to produce the output of a larger model. Neural Network Pruning is performed at multiple levels; (i) weight pruning~\cite{mozer1989using, han2015learning, han2015deep} removes per parameter basis while (ii) neuron/channel~\cite{wen2016learning, lebedev2016fast} pruning removes per neuron or channel basis and (iii) block/group~\cite{gordon2018morphnet, leclerc2018smallify} pruning removes per a block of network such as residual block or sub-network.  

% - Why pruning, why is it important.

Weight pruning generally achieves a very high pruning ratio getting similar performance only with a few percentages of the parameters. This allows a high compression of the network and also accelerates the network on specialized hardware and CPUs. However, weight pruning in a defined format such as N:M block-sparse helps in improving the performance on GPUs~\cite{liu2023ten}. Pruning network at the level of neurons or channels helps in reducing the parameters with similar performance, however, the pruning ratio is not that high. Furthermore, pruning at the level of blocks helps to reduce the complexity of the network and creates a smaller network for faster inference. All these methods can be applied to the same model as well. Furthermore, pruning and addition of neurons can bring dynamic behaviour of decreasing and increasing network capacity which has found its application in Continual or Incremental Learning settings as well as Neural Architecture Search~\cite{gordon2018morphnet, zhang2020regularize, dai2020incremental, sapkota2022noisy}.

% - Why neuron level pruning, why global importance.

In this work, we are particularly interested in neuron level pruning. Apart from the benefit of reduced parameter, memory and compution, neuron/channel level pruning is more similar to a biological formulation where the neurons are the basic unit of computation. Furthermore, the number of neurons in a neural network is small compared to the number of connections and can easily be pruned by measuring the global importance~\cite{lecun1989optimal, hassibi1993optimal, molchanov2016pruning, lee2018snip, yu2018nisp}. We focus on the global importance as it removes the need to inject bias about the number of neurons to prune in each layer. This can simplify our problem to remove less significant neurons globally which allows us to extend it to unorganized networks other than layered formulation. However, in this work, we focus only on layer-wise or block-wise architectures such as ResNet and VGG.

% - What is the observation from previous experiments/papers and what are the problems ?

Previous works show that global importance estimation can be computed using one or all of forward/activation~\cite{hu2016network}, parameter/weight~\cite{han2015learning} or backward/gradient~\cite{wang2020picking, lubana2020gradient, evci2022gradient} signals. Some of the previous techniques use Feature Importance propagation ~\cite{yu2018nisp} or Gradient propagation~\cite{lee2018snip} to find the neuron importance. Taylor First Order approximation based methods use both activation and gradient information for pruning~\cite{molchanov2016pruning, molchanov2019importance}. There are also works that improve pruning with Taylor Second Order approximations~\cite{singh2020woodfisher, yu2022combinatorial, chen2022network}. Although there are methods using forward and backward signals for pruning at initialization~\cite{wang2020picking}, we limit our experiment to the pruning of trained models for a given number of neurons.

% - What we do in this paper: methods and observation.

In this work, we derive a similar importance metric to Taylor First Order (Taylor-FO) approximations~\cite{molchanov2016pruning, molchanov2019importance} but from heuristics combining both forward and backward signals. The forward signal, namely the activation of the neuron, and the backward signal, the gradient or the random gradient of the output. We also compare the pruning accuracy with a different number of data samples and find that our method performs better than the previous works in most settings. Furthermore, our method of using the random gradient signal on the last layer and gradient normalization is also applicable to previous methods, showing that our approach improves performance on previous methods as well.

\begin{figure*}[h]
    \centering
    \includegraphics[width=0.99\linewidth]{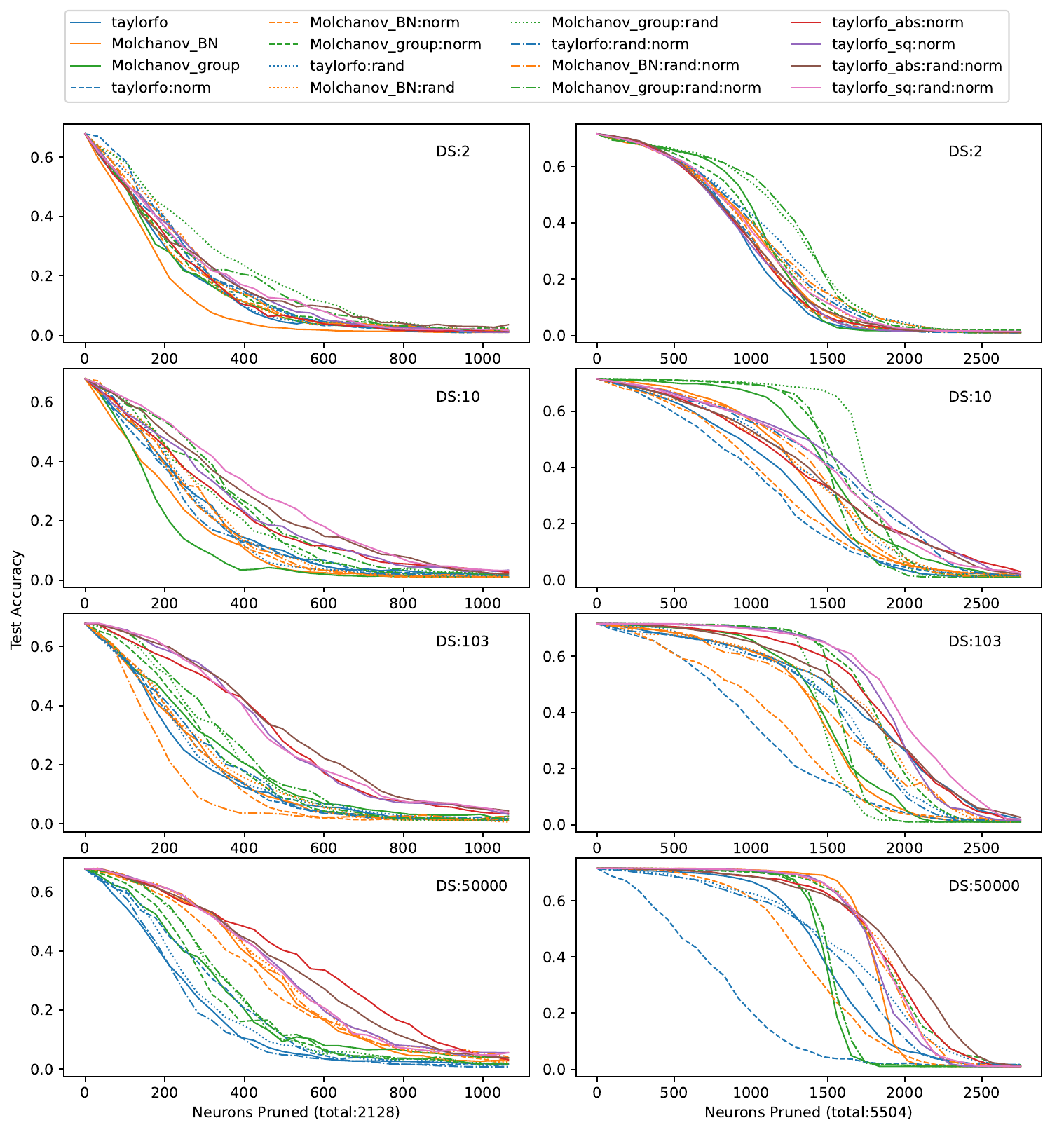}
    \caption{The number of neurons pruned vs Accuracy plot for (left) ResNet-54 and (right) VGG19 for various data sizes (DS). Methods: \texttt{taylorfo}~\cite{molchanov2016pruning}, \texttt{Molchanov\_BN} and \texttt{Molchanov\_group}~\cite{molchanov2019importance} are the baselines while \texttt{taylorfo\_sq} and \texttt{taylorfo\_abs} are \emph{our methods}. \texttt{rand} represents use of random gradient and \texttt{norm} represents use of gradient magnitude normalization.}
    \label{fig:c100_resnet}
\end{figure*}

\section{Motivation and Our Method}

% - The limitation of existing methods.

Previous works on global importance based post-training pruning of neurons focus on using forward and backward signals. Since most of these methods are based on Taylor approximation of the change in loss after the removal of a neuron or group of parameters~\cite{}, these methods require input and target value for computing the importance.

% - Our approach at the same problem from different angle.

We tackle the problem of pruning from the perspective of overall function output without considering the loss.

\textbf{Forward Signal:} The forward signal is generally given by the pre-activation ($x_i$). If a pre-activation is zero, then it has no impact on the output of the function, i.e. the output deviation with respect to the removal of the neuron is zero. If the incoming connection of a neuron is zero-weights, then the neuron can be simply removed, i.e. it has no significance. If the incoming connection is non-zero then the neuron has significance. Forward signal takes into consideration how data affects a particular neuron.

\textbf{Backward Signal:} The backward signal is generally given by back-propagating the loss. If the outgoing connection of the neuron is zeros, then the neuron has no significance to the function even if it has positive activation. The gradient($\delta x_i$) provides us with information on how the function or loss will change if the neuron is removed.

\textbf{Importance Metric:} Combining the forward and backward signal we can get the influence of the neuron on the loss or the function for given data. Hence, the importance metric ($I_i$) of each neuron ($n_i$) for dataset of size $M$ is given by $I_i = \frac{1}{M}\sum_{n=1}^{M}{x_i.\delta x_i}$, where $x_i$ is the pre-activation and $\delta x_i$ is its gradient. It fulfils the criterion that importance should be low if incoming or outgoing connections are zeros and higher otherwise.

\textbf{Problem 1:} This importance metric is similar to Taylor-FO~\cite{molchanov2016pruning}. However, analysing this, we find that this method produces low importance if the gradient is negative, which to our application is a problem as the function will be changed significantly, even if it lowers the loss. Hence, we make the importance positive either by using the absolute value or the squared value. This gives rise to two of our methods namely:

\textbf{Taylor-FO-abs :} $I_i = \frac{1}{M}\sum_{n=1}^{M}{|x_i.\delta x_i|}$  \hspace{4mm}
\textbf{Taylor-FO-sq :} $I_i = \frac{1}{M}\sum_{n=1}^{M}{(x_i.\delta x_i)^2}$

\textbf{Problem 2:} The goal of neuron removal is to make the least change in the loss value, which requires the label to calculate the loss. However, according to our intuition, the goal should be to make the least change in the output of the function. Having the least change in the function implicitly fulfils the objective of previous methods to have the least change in the loss. This removes the need to have labels for pruning. Furthermore, viewing from the backward signal, the slope/gradient of the output towards the correct label should not be the requirement, as data points getting the same output as the target will not produce any gradient and hence deemed insignificant by previous methods. 

\begin{figure*}[t]
    \centering
    \includegraphics[width=0.99\linewidth]{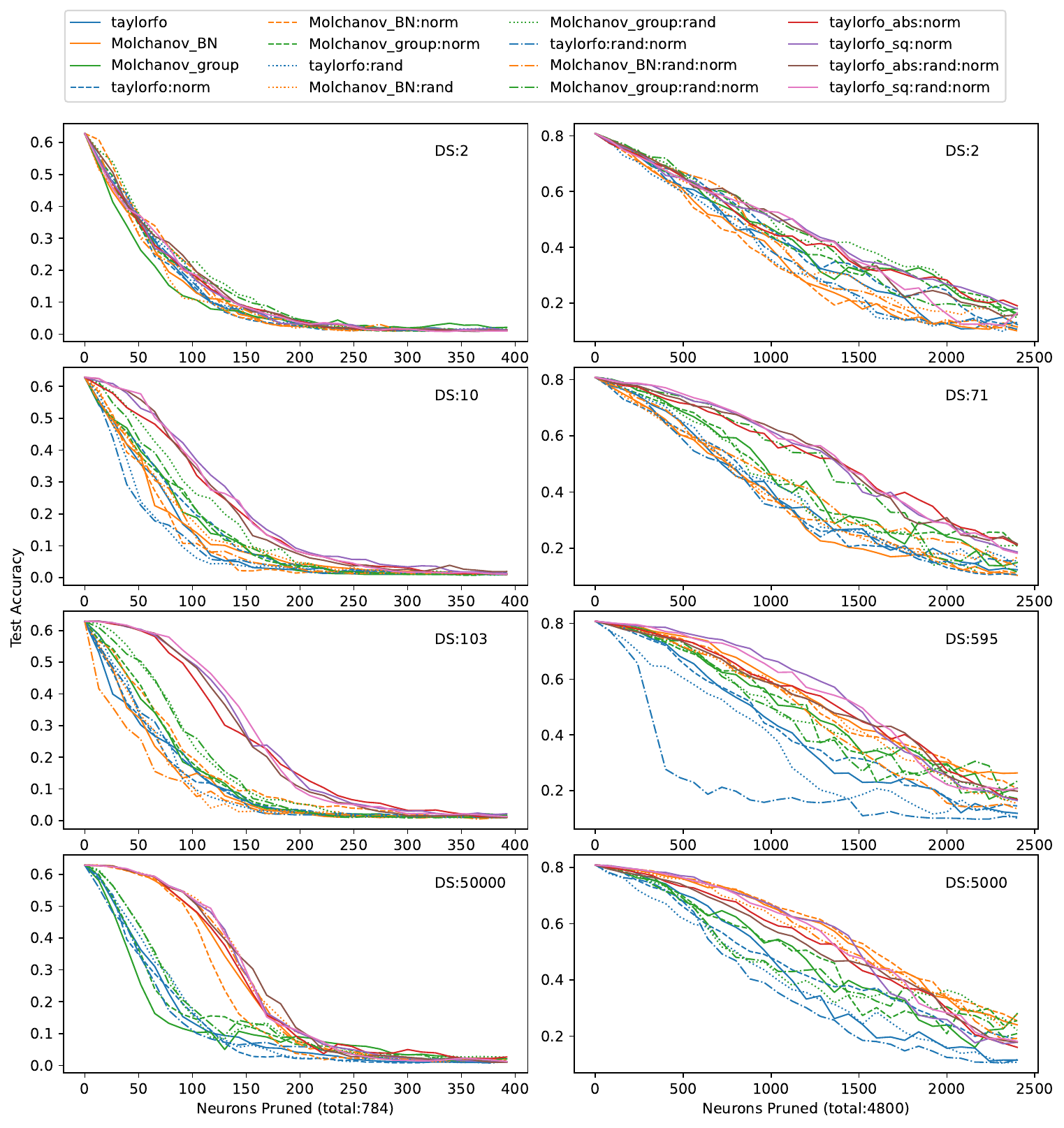}
    \caption{
    The number of neurons pruned vs Accuracy plot for (left) ResNet-20 on CIFAR-100 and (right) ReNet-18 on STL-10 dataset for various data sizes (DS). Methods: 
    \texttt{taylorfo}~\cite{molchanov2016pruning},
    \texttt{Molchanov\_BN} and \texttt{Molchanov\_group}~\cite{molchanov2019importance} are the baselines while \texttt{taylorfo\_sq} and \texttt{taylorfo\_abs} are \emph{our methods}. \texttt{rand} represents use of random gradient and \texttt{norm} represents use of gradient magnitude normalization.
    }
    \label{fig:stl10_resnet}
\end{figure*}

Alternatively, we test the hypothesis that any \emph{random gradient} should work fine, and the gradient should be \emph{normalized} to the same magnitude (of $1$). Doing so makes the contribution of each data point equal for computing the importance. We find that the use of random gradient also produces similar pruning performance as shown in Experiment Section Figure~\ref{fig:c100_resnet}and~\ref{fig:stl10_resnet} supporting our hypothesis.

% - What our method unlocks and its possible application.

Our method of using random gradient can be applied to all the previous methods using gradient signal from output to compute the importance. The implication can be useful in the setting of unsupervised pruning, where a network can be pruned to target datasets without labels. The application of this method can be significant for settings such as Reinforcement Learning, where the input data is abundant but the label is scarce.

\textbf{Similarity to Molchanov-BN pruning:}
Our method is similar to Molchanov - Batch Norm pruning~\cite{molchanov2019importance} as both of our methods perform neuron/channel level pruning using forward and backward signals.

Molchanov-BN has Importance given by $I_i = \sum(\gamma_i.\delta\gamma_i + \beta_i.\delta\beta_i)^2$, where $\gamma$, $\beta_i$ represents the scaler and bias terms of BN and $\delta \gamma$ and $\delta \beta$ represents their gradient respectively. If we consider $\beta = 0$ then $I_i = \sum(\gamma_i.\delta\gamma_i)^2$. If we consider the input of batch-norm to be $x_i$ and output after scaling to be $y_i = x_i.\gamma_i$ then the gradient $\delta \gamma_i = \delta y_i . x_i$ and the overall importance is $I_i = \sum(\gamma_i.\delta y_i . x_i)^2$.

In our case, we take the importance $I_i = \sum(x_i.\delta x_i)^2$. Considering the value with respect to $\delta y_i$, we get our importance to be $I_i = \sum x_i . \gamma . \delta y_i$. Which differs only by the constant shift term $\beta$. It turns out that our method produces better accuracy for ResNet style architecture. However, in VGG architecture, our pruning method performs better than Molanchonov-BN and is competitive or better than Molanchonov-WeightGroup when tested in various settings.

% - Initial experiments to verify that our method works.

% To check if our newly proposed two methods alongside use of random gradient works or not, we experiment on MNIST for global importance based pruning and find that our methods are comparable to Taylor. Our random gradient method proves to imporve the pruning accuracy for previous methods as well as our method.

% \section{Pruning with and without labels}
% - Explain experiments on Resnets/VGG for different methods

% \section{Pruning with varying dataset size}
% - Explain experiments with different number of training samples (N). Compare it with random gradient for (N) samples as well as full samples.

\section{Experiments}

\textbf{Procedure:} First, we compute the importance score for each neuron/channel on a given dataset. Secondly, we prune the $P$ least important neurons of total $N$ by importance metric ($I$) given by different methods. We measure the resulting accuracy and plot with the number of neurons pruned. We test for Taylor method~\cite{molchanov2016pruning, molchanov2019importance}, and on two of our methods. The pruning is performed on the training dataset and accuracy is measured on the test dataset.

Furthermore, to test the performance of different methods on different dataset sizes, we test for $D$ data points of total $M$ data points on all the datasets and architectures. The total dataset size for CIFAR-100 is 50K and for STL-10 is 5K. 

\textbf{CIFAR-100: }
We test the pruning performance of different methods on CIFAR-100~\cite{krizhevsky2009learning} dataset for dataset size $D \in [ 2, 10, 103, 50K]$ as shown in the Figure~\ref{fig:c100_resnet}~\ref{fig:stl10_resnet}. We test the model for Resnet-20, ResNet-56 and VGG-19.

\textbf{STL-10:} ResNet-18 model used on STL-10~\cite{coates2011analysis} dataset is finetuned on Imagenet pretrained model. We test the pruning performance of different methods on the STL-10 dataset for dataset size $D \in [2, 71, 595, 5K]$ as shown in Figure~\ref{fig:stl10_resnet}.

\section{Conclusion}

In this paper, we improve on previously proposed Taylor First Order based pruning methods~\cite{molchanov2019importance} by using random gradients and by normalizing the gradients. We show that our techniques improve on the previous methods as well as our own variation. The knowledge that these methods allow for pruning with random gradient backpropagation is the main contribution of our work.



\clearpage
\bibliography{iclr2023_conference}
\bibliographystyle{iclr2023_conference}

% \appendix
% \section{Appendix}

\end{document}